\documentclass[10pt,twocolumn,letterpaper]{article}

\usepackage{iccv}
\usepackage{times}
\usepackage{epsfig}
\usepackage{graphicx}
\usepackage{amsmath}
\usepackage{amssymb}

\usepackage{xspace}
\usepackage{subcaption}
\usepackage{multirow}
\usepackage[labelfont=bf]{caption}

\usepackage[pagebackref=true,breaklinks=true,letterpaper=true,colorlinks,bookmarks=false]{hyperref}

\usepackage[inline]{enumitem}
\usepackage{makecell}

\iccvfinalcopy 

\def\iccvPaperID{581} 

\graphicspath{{./figures/}}

\newcommand{\architecture}[1]{\emph{#1}}
\newcommand{\arch}[1]{\emph{#1}}
\newcommand{\mrcell}[1]{\multicolumn{1}{l|}{\begin{tabular}[l]{@{}l@{}} #1 \end{tabular}}}
\newcommand{\lmrcell}[1]{\multicolumn{1}{|l|}{\begin{tabular}[l]{@{}l@{}} #1 \end{tabular}}}

\ificcvfinal\fi

\begin{document}

\title{Back to Simplicity: How to Train Accurate BNNs from Scratch?}

\author{Joseph Bethge\thanks{Authors contributed equally} , Haojin Yang\footnotemark[1] , Marvin Bornstein, Christoph Meinel  \\
Hasso Plattner Institute, University of Potsdam, Germany\\
{\tt\small \{joseph.bethge,haojin.yang,meinel\}@hpi.de}, {\tt\small \{marvin.bornstein\}@student.hpi.de}
}

\maketitle

\begin{abstract}


\noindent
Binary Neural Networks (BNNs) show promising progress in reducing computational and memory costs but suffer from substantial accuracy degradation compared to their real-valued counterparts on large-scale datasets, e.g., ImageNet.
Previous work mainly focused on reducing quantization errors of weights and activations, whereby a series of approximation methods and sophisticated training tricks have been proposed.
In this work, we make several observations that challenge conventional wisdom. 
We revisit some commonly used techniques, such as scaling factors and custom gradients, and show that these methods are not crucial in training well-performing BNNs.
On the contrary, we suggest several design principles for BNNs based on the insights learned and 
demonstrate that highly accurate BNNs can be trained from scratch with a simple training strategy.
We propose a new BNN architecture BinaryDenseNet, which significantly surpasses all existing 1-bit CNNs on ImageNet without tricks.
In our experiments, BinaryDenseNet achieves $18.6\%$ and $7.6\%$ relative improvement over the well-known XNOR-Network and the current state-of-the-art Bi-Real Net in terms of top-1 accuracy on ImageNet, respectively.
\url{https://github.com/hpi-xnor/BMXNet-v2}
\end{abstract}


\begin{table*}[]
\centering
\caption{
A general comparison of the most related methods to this work.
Essential characteristics such as value space of inputs and weights, numbers of multiply-accumulate operations (MACs), numbers of binary operations, theoretical speedup rate and operation types, are depicted.
The results are based on a \textbf{single} quantized convolution layer from each work.
$\beta$ and $\alpha$ denote the full-precision scaling factor used in proper methods, whilst m, n, k denote the dimension of weight ($W \in \mathbb{R}^{n \times k}$) and input ($I$ $\in$ $\mathbb{R}^{k \times m}$).
The table is adapted from \cite{Wan_2018_ECCV}.
} 
\begin{tabular}{|c|c|c|c|c|c|c|}
\hline
 Methods     &    Inputs           & 		Weights 	& 	MACs  &  Binary Operations  &   
 Speedup  	& 	  Operations \\ \hline
 Full-precision     &     $\mathbb{R}$         & 		$\mathbb{R}$ 	&  n$\times$m$\times$k  &  0  &   1$\times$  	& 	  mul,add \\ \hline

 BC \cite{courbariaux2015binaryconnect}    &     $\mathbb{R}$         & 		$\{-1,1\}$ 	&  n$\times$m$\times$k  &  0  &   $\sim$ 2$\times$  	& 	  sign,add \\ \hline

 BWN \cite{Rastegari2016}    &     $\mathbb{R}$         & 		$\{-\alpha,\alpha\}$ 	&  n$\times$m$\times$k  &  0  &   $\sim$ 2$\times$  	& 	  sign,add \\ \hline

 TTQ \cite{zhu2016trained}    &     $\mathbb{R}$         & 		$\{-\alpha^n,0,\alpha^p\}$ 	&  n$\times$m$\times$k  &  0  &   $\sim$ 2$\times$  	& 	  sign,add \\ \hline

 DoReFa \cite{Zhou2016}    &     $\{0,1\}\times$4         & 		$\{0,\alpha\}$ 	&  n$\times$k  &  8$\times$n$\times$m$\times$k  &  $\sim$ 15$\times$  	& 	  and,bitcount \\ \hline

 HORQ \cite{li2017performance}    &     $\{-\beta,\beta\}\times$2         & 		$\{-\alpha,\alpha\}$ 	&  4$\times$n$\times$m  &  4$\times$n$\times$m$\times$k  &  $\sim$ 29$\times$  	& 	  xor,bitcount \\ \hline

 TBN \cite{Wan_2018_ECCV}    &     $\{-1,0,1\}$         & 		$\{-\alpha,\alpha\}$ 	&  n$\times$m  &  3$\times$n$\times$m$\times$k  &  $\sim$ 40$\times$  	& 	  and,xor,bitcount \\ \hline

 XNOR \cite{Rastegari2016}    &     $\{-\beta,\beta\}$       & 		$\{-\alpha,\alpha\}$ 	&  2$\times$n$\times$m  &  2$\times$n$\times$m$\times$k  &  $\sim$ 58$\times$  	& 	  xor,bitcount \\ \hline

 BNN \cite{Courbariaux2016}    &     $\{-1,1\}$         & 		$\{-1,1\}$ 	&  0  &  2$\times$n$\times$m$\times$k  &  $\sim$ 64$\times$  	& 	  xor,bitcount \\ \hline

 Bi-Real \cite{Liu_2018_ECCV}    &     $\{-1,1\}$         & 		$\{-1,1\}$ 	&  0  &  2$\times$n$\times$m$\times$k  &  $\sim$ 64$\times$  	& 	  xor,bitcount \\ \hline

 \textbf{Ours}    &     $\{-1,1\}$         & 		$\{-1,1\}$ 	&  0  &  2$\times$n$\times$m$\times$k  &  $\sim$ 64$\times$  	& 	  xor,bitcount \\ \hline

\end{tabular}
\label{tab:related-methods}
\end{table*}

\section{Introduction}
\label{sec:intro}

\noindent
Convolutional Neural Networks have achieved state-of-the-art on a variety of tasks related to computer vision, for example, classification \cite{cifar10}, detection \cite{girshick2014rich}, and text recognition \cite{jaderberg_2014_eccv}.
By reducing memory footprint and accelerating inference, there are two main approaches which allow for the execution of neural networks on devices with low computational power, \eg mobile or embedded devices:
On the one hand, information in a CNN can be compressed through compact network design.
Such methods use full-precision floating point numbers as weights, but reduce the total number of parameters and operations through clever network design, while minimizing loss of accuracy, \eg, \arch{SqueezeNet} \cite{Iandola2016}, \arch{MobileNets} \cite{Howard2017}, and \arch{ShuffleNet} \cite{Zhang2017}.
On the other hand, information can be compressed by avoiding the common usage of full-precision floating point weights and activations, which use 32 bits of storage.
Instead, quantized floating-point numbers with lower precision (\eg 4 bit of storage) \cite{Zhou2016} or even binary (1 bit of storage) weights and activations \cite{Courbariaux2016,lin2017towards,Liu_2018_ECCV,Rastegari2016} are used in these approaches.
A BNN achieves up to $32\times$ memory saving and $58\times$ speedup on CPUs by representing both weights and activations with binary values \cite{Rastegari2016}.
Furthermore, computationally efficient, bitwise operations such as $\mathrm{xnor}$ and $\mathrm{bitcount}$ can be applied for convolution computation instead of arithmetical operations.
Despite the essential advantages in efficiency and memory saving, BNNs still suffer from the noticeable accuracy degradation that prevents their practical usage.
To improve the accuracy of BNNs, previous approaches mainly focused on reducing quantization errors by using complicated approximation methods and training tricks, such as scaling factors \cite{Rastegari2016}, multiple weight/activation bases \cite{lin2017towards}, fine-tuning a full-precision model, multi-stage pre-training, or custom gradients \cite{Liu_2018_ECCV}.
These work applied well-known real-valued network architectures such as \arch{AlexNet}, \arch{GoogLeNet} or \arch{ResNet} to BNNs without thorough explanation or experiments on the design choices.
However, they don't answer the simple yet essential question: \arch{Are those real-valued network architectures seamlessly suitable for BNNs?}
Therefore, appropriate network structures for BNNs should be adequately explored.

In this work, we first revisit some commonly used techniques in BNNs.
Surprisingly, our observations do not match conventional wisdom.
We found that most of these techniques are \emph{not} necessary to reach state-of-the-art performance. 
On the contrary, we show that highly accurate BNNs can be trained from scratch by ``simply'' maintaining rich information flow within the network.
We present how increasing the number of shortcut connections improves the accuracy of BNNs significantly
and demonstrate this by designing a new BNN architecture \arch{BinaryDenseNet}.
Without bells and whistles, \arch{BinaryDenseNet} reaches state-of-the-art by using standard training strategy which is much more efficient than previous approaches.

Summarized, our contributions in this paper are:
\vspace{-0.1cm}
\begin{itemize}
	\itemsep0em
	\item We show that highly accurate binary models can be trained by using standard training strategy, which challenges 		conventional wisdom.
		We analyze why applying common techniques (as \eg, scaling methods, custom gradient, and fine-tuning a full-precision model) is ineffective when training from scratch and provide empirical proof.
	\item We suggest several general design principles for BNNs and 
		further propose a new BNN architecture \arch{BinaryDenseNet}, which significantly surpasses all existing 1-bit CNNs for image classification without tricks.		
	\item To guarantee the reproducibility, we contribute to an open source framework for BNN/quantized NN.
		We share codes, models implemented in this paper for classification and object detection.
		Additionally, we implemented the most influential BNNs including \cite{Courbariaux2016,lin2017towards,Liu_2018_ECCV,Rastegari2016,Zhou2016} to facilitate follow-up studies.
\end{itemize}
\vspace{-0.1cm}

The rest of the paper is organized as follows: We describe related work in \autoref{sec:related_work}.
We revisit common techniques used in BNNs in \autoref{sec:revisit}.
\autoref{sec:approach} and \ref{sec:main_results} present our approach and the main result.


\section{Related work}
\label{sec:related_work}
\noindent
In this section, we roughly divide the recent efforts for binarization and compression into three categories:
\begin{enumerate*}[label=(\roman*)]
\item compact network design,
\item networks with quantized weights,
\item and networks with quantized weights and activations.
\end{enumerate*}

\noindent
\textbf{Compact Network Design.} 
This sort of methods use full-precision floating point numbers as weights, but reduce the total number of parameters and operations through compact network design, while minimizing loss of accuracy.
The commonly used techniques include replacing a large portion of 3$\times$3 filters with smaller 1$\times$1 filters \cite{Iandola2016};
Using depth-wise separable convolution to reduce operations \cite{Howard2017};
Utilizing channel shuffling to achieve group convolutions in addition to depth-wise convolution \cite{Zhang2017}.
These approaches still require GPU hardware for efficient training and inference.
A strategy to accelerate the computation of all these methods for CPUs has yet to be developed.

\noindent
\textbf{Quantized Weights and Real-valued Activations.}
Recent efforts from this category, for instance, include \emph{BinaryConnect} (BC) \cite{courbariaux2015binaryconnect}, \emph{Binary Weight Network} (BWN) \cite{Rastegari2016}, and \emph{Trained Ternary Quantization} (TTQ) \cite{zhu2016trained}.
In these work, network weights are quantized to lower precision or even binary.
Thus, considerable memory saving with relatively little accuracy loss has been achieved.
But, no noteworthy acceleration can be obtained due to the real-valued inputs.

\noindent
\textbf{Quantized Weights and Activations.}
On the contrary, approaches adopting quantized weights and activations can achieve both compression and acceleration.
Remarkable attempts include \emph{DoReFa-Net} \cite{Zhou2016}, \emph{High-Order Residual Quantization} (HORQ) \cite{li2017performance} and SYQ \cite{faraone2018syq}, which reported promising results on ImageNet \cite{imagenet_cvpr09} with 1-bit weights and multi-bits activations.

\noindent
\textbf{Binary Weights and Activations.}
BNN is the extreme case of quantization, where both weights and activations are binary.
Hubara \etal proposed \emph{Binarized Neural Network} (BNN) \cite{Courbariaux2016},
where weights and activations are restricted to +1 and -1.
They provide efficient calculation methods for the equivalent of matrix multiplication by using $\mathrm{xnor}$ and $\mathrm{bitcount}$ operations.
\emph{XNOR-Net} \cite{Rastegari2016} improved the performance of BNNs by introducing a channel-wise scaling factor to reduce the approximation error of full-precision parameters.
\emph{ABC-Nets} \cite{lin2017towards} used multiple weight bases and activation bases to approximate their full-precision counterparts.
Despite the promising accuracy improvement, the significant growth of weight and activation copies offsets the memory saving and speedup of BNNs.
Wang \etal. \cite{Wan_2018_ECCV} attempted to use binary weights and ternary activations in their \emph{Ternary-Binary Network} (TBN). 
They achieved a certain degree of accuracy improvement with more operations compared to fully binary models.
In \emph{Bi-Real Net}, Liu \etal \cite{Liu_2018_ECCV} proposed several modifications on \arch{ResNet}.
They achieved state-of-the-art accuracy by applying an extremely sophisticated training strategy that consists of full-precision pre-training, multi-step initialization (ReLU$\rightarrow$leaky clip$\rightarrow$clip \cite{ZechunLiu2018}), and custom gradients.

\autoref{tab:related-methods} gives a thorough overview of the recent efforts in this research domain.
We can see that our work follows the most straightforward binarization strategy as BNN \cite{Courbariaux2016}, that achieves the highest theoretical speedup rate and the highest compression ratio.
Furthermore, we directly train a binary network from scratch by adopting a simple yet effective strategy.


\section{Study on Common Techniques}
\label{sec:revisit}

\begin{table}[]
\caption{
    The influence of using scaling, a full-precision downsampling convolution, and the $\mathrm{approxsign}$ function on the CIFAR-10 dataset based on a binary \architecture{ResNetE18}.
    Using $\mathrm{approxsign}$ instead of $\mathrm{sign}$ slightly boosts accuracy, but only if training a model with scaling factors. 
}
\begin{tabular}{|l|l|l|l|}
\hline
\lmrcell{Use \\ scaling \\ of \cite{Rastegari2016}} & \mrcell{Downsampl. \\ convolution} & \mrcell{Use \\ $\mathrm{approxsign}$ \\ of \cite{Liu_2018_ECCV}} & \mrcell{Accuracy \\ Top1/Top5} \\ \hline
\multirow{4}{*}{no}  & \multirow{2}{*}{binary}         & yes         & 84.9\%/99.3\%                   \\ \cline{3-4} 
                     &                                 & no          & 87.2\%/\textbf{99.5\%}          \\ \cline{2-4} 
                     & \multirow{2}{*}{full-precision} & yes         & 86.1\%/99.4\%                   \\ \cline{3-4} 
                     &                                 & no          & \textbf{87.6\%}/\textbf{99.5\%} \\ \hline
\multirow{4}{*}{yes} & \multirow{2}{*}{binary}         & yes         & 84.2\%/99.2\%                   \\ \cline{3-4} 
                     &                                 & no          & 83.6\%/99.2\%                   \\ \cline{2-4} 
                     & \multirow{2}{*}{full-precision} & yes         & 84.4\%/99.3\%                   \\ \cline{3-4} 
                     &                                 & no          & 84.7\%/99.2\%                   \\ \hline
\end{tabular}
\label{tab:cifar-scaling-downsampling-approxsign}
\end{table}

\noindent
In this section, to ease the understanding, we first provide a brief overview of the major implementation principles of a binary layer (see supplementary materials for more details).
We then revisit three commonly used techniques in BNNs: scaling factors \cite{Rastegari2016,Zhou2016,Wan_2018_ECCV,Tang2017,li2017performance,zhu2016trained,lin2017towards}, full-precision pre-training \cite{Zhou2016,Liu_2018_ECCV}, and $\mathrm{approxsign}$ function \cite{Liu_2018_ECCV}.
We didn't observe accuracy gain as expected.
We analyze why these techniques are not as effective as previously presented when training from scratch and provide empirical proof.
The finding from this study motivates us to explore more effective solutions for training accurate BNNs.

\subsection{Implementation of Binary Layers}
\label{sec:implementation-details}

\noindent
We apply the sign function for binary activation, thus transforming floating-point values into binary values:
\begin{equation}
    \mathrm{sign}(x) = \begin{cases} 
    +1 ~\text{if}~ x \geq 0, \\
    -1 ~\text{otherwise}.
    \end{cases}
\end{equation}
The implementation uses a Straight-Through Estimator (STE) \cite{BengioLC13_STE} with the addition, that it cancels the gradients, when the inputs get too large, as proposed by Hubara \etal \cite{Courbariaux2016}.
Let $c$ denote the objective function, $r_i$ be a real number input, and $r_o\in\{-1,+1\}$ a binary output.
Furthermore, $t_\mathrm{clip}$ is the threshold for clipping gradients, which was set to $t_\mathrm{clip}=1$ in previous works~\cite{Zhou2016,Courbariaux2016}.
Then, the resulting STE is:
\begin{align}
\label{eq:STE-forward}
    \text{Forward:}& ~r_o=\mathrm{sign}(r_i)~. \\
\label{eq:STE-backward}
    \text{Backward:}& ~\frac{\partial c}{\partial r_i}=\frac{\partial c}{\partial r_o}1_{|r_i|\leq t_\mathrm{clip}}~.
\end{align}

\subsection{Scaling Methods}
\label{sec:scaling}

\begin{figure}[]
    \centering
    \includegraphics[width=\linewidth]{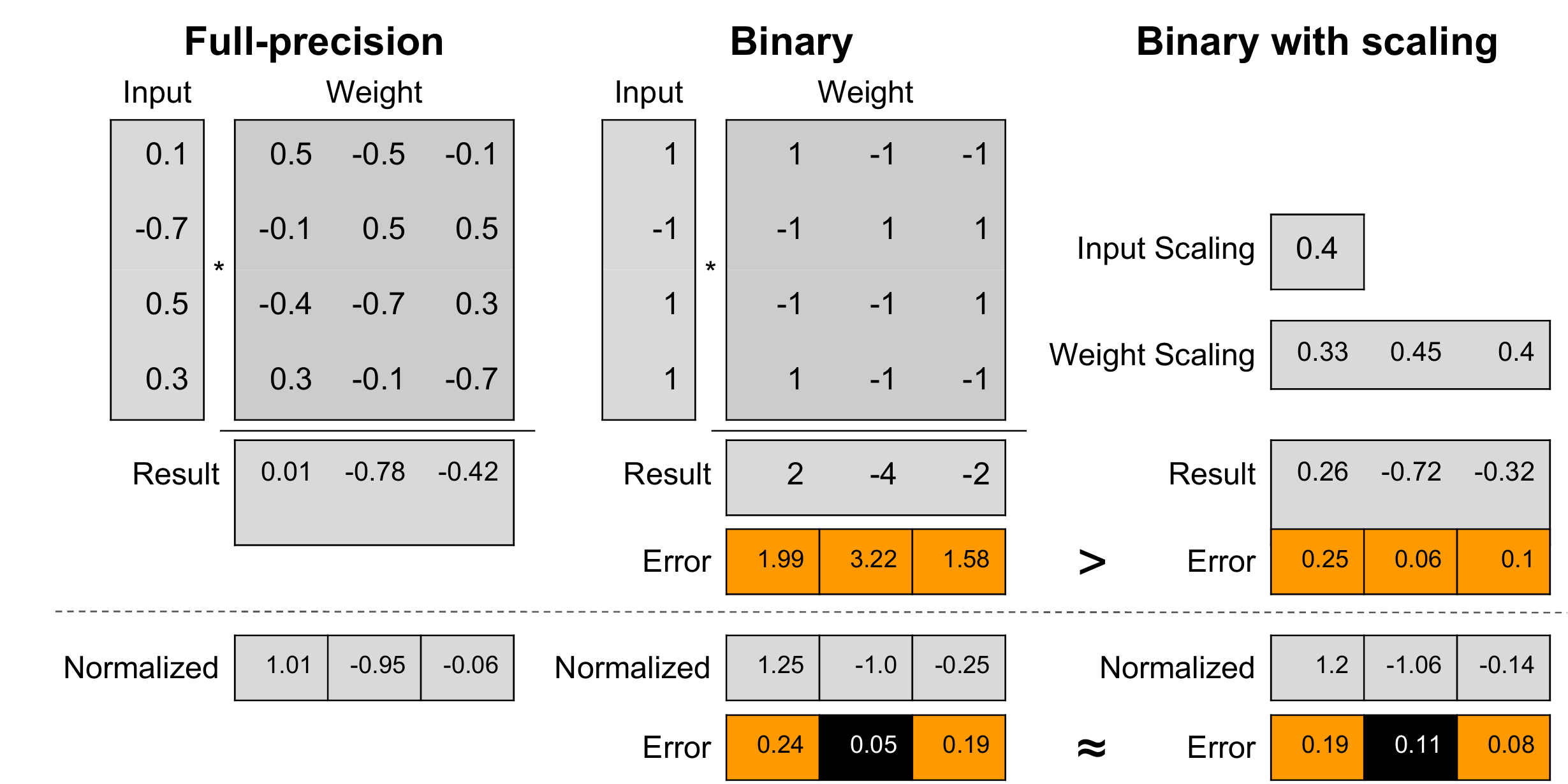}
    \caption{
        An exemplary implementation shows that normalization minimizes the difference between a binary convolution with scaling (right column) and one without (middle column). 
        In the top row, the columns from left to right respectively demonstrate the \emph{gemm} results of full-precision, binary, and binary with scaling.
        The bottom row shows their results after normalization.
        Errors are the absolute difference between full-precision and binary results.
        The results indicate that normalization dilutes the effect of scaling.
    }
    \label{fig:scaling}
\end{figure}

\noindent
Binarization will always introduce an approximation error compared to a full-precision signal.
In their analysis, Zhou \etal \cite{Zhou2017} show that this error linearly degrades the accuracy of a CNN.

Consequently, Rastegari \etal \cite{Rastegari2016} propose to scale the output of the binary convolution by the average absolute weight value per channel ($\alpha$) and average absolut activation over all input channels ($K$).
\begin{align}
    \mathbf{x} \ast \mathbf{w} &\approx \mathrm{binconv}(\mathrm{sign}(x), \mathrm{sign}(w)) \cdot K \cdot \alpha
\end{align}

The scaling factors should help binary convolutions to increase the value range.
Producing results closer to those of full-precision convolutions and reducing the approximation error.
However, these different scaling values influence specific output channels of the convolution.
Therefore, a BatchNorm \cite{ioffe2015batch} layer directly after the convolution (which is used in all modern architectures) theoretically minimizes the difference between a binary convolution with scaling and one without.
Thus, we hypothesize that learning a useful scaling factor is made inherently difficult by BatchNorm layers.
\autoref{fig:scaling} demonstrates an exemplary implementation of our hypothesis.

We empirically evaluated the influence of scaling factors (as proposed by Rastegari \etal \cite{Rastegari2016}) on the accuracy of our trained models based on the binary \architecture{ResNetE} architecture (see \autoref{sec:resnetE}).
First, the results of our CIFAR-10 \cite{cifar10} experiments verify our hypothesis, that applying scaling when training a model from scratch does not lead to better accuracy (see \autoref{tab:cifar-scaling-downsampling-approxsign}).
All models show a decrease in accuracy between 0.7\% and 3.6\% when applying scaling factors.
Secondly, we evaluated the influence of scaling for the ImageNet dataset (see \autoref{tab:imagenet-scaling-downsampling-approxsign}).
The result is similar, scaling reduces model accuracy ranging from 1.0\% to 1.7\%.
We conclude that the BatchNorm layers following each convolution layer absorb the effect of the scaling factors.
To avoid the additional computational and memory costs, we don't use scaling factors in the rest of the paper.

\begin{table}[]
\caption{
    The influence of using scaling, a full-precision downsampling convolution, and the $\mathrm{approxsign}$ function on the ImageNet dataset based on a binary \architecture{ResNetE18}.
}
\begin{tabular}{|l|l|l|l|}
\hline
\lmrcell{Use \\ scaling \\ of \cite{Rastegari2016}} & \mrcell{Downsampl. \\ convolution} & \mrcell{Use \\ $\mathrm{approxsign}$ \\ of \cite{Liu_2018_ECCV}} & \mrcell{Accuracy \\ Top1/Top5} \\ \hline
\multirow{4}{*}{no}  & \multirow{2}{*}{binary}         & yes         & 54.3\%/77.6\%           \\ \cline{3-4}
                     &                                 & no          & 54.5\%/77.8\%           \\ \cline{2-4}
                     & \multirow{2}{*}{full-precision} & yes         & 56.6\%/79.3\%  \\ \cline{3-4}
                     &                                 & no          & \textbf{58.1\%}/\textbf{80.6\%}  \\ \hline
\multirow{4}{*}{yes} & \multirow{2}{*}{binary}         & yes         & 53.3\%/76.4\%           \\ \cline{3-4} 
                     &                                 & no          & 52.7\%/76.1\%           \\ \cline{2-4} 
                     & \multirow{2}{*}{full-precision} & yes         & 55.3\%/78.3\%           \\ \cline{3-4} 
                     &                                 & no          & 55.6\%/78.4\%           \\ \hline
\end{tabular}
\label{tab:imagenet-scaling-downsampling-approxsign}
\end{table}

\begin{table*}[]
\centering
\caption{
Comparison of our binary \arch{ResNetE18} model to state-of-the-art binary models using ResNet18 on the ImageNet dataset.
The top-1 and top-5 validation accuracy are reported.
For the sake of fairness we use the ABC-Net result with 1 weight base and 1 activation base in this table.
} 
\begin{small}
\begin{tabular}{|l|c|c|c|c|c|c|c|}
\hline
 \thead{Downsampl.\\ convolution} & Size & Our result & Bi-Real \cite{Liu_2018_ECCV} & TBN \cite{Wan_2018_ECCV} & HORQ \cite{li2017performance} & XNOR \cite{Rastegari2016} & ABC-Net (1/1) \cite{lin2017towards}   \\ \hline
 
 full-precision & 4.0 MB & \textbf{58.1\%}/\textbf{80.6\%} & 56.4\%/79.5\%  & 55.6\%/74.2\% & 55.9\%/78.9\%  & 51.2\%/73.2\%  & n/a \\ \hline
 
 binary & 3.4 MB & \textbf{54.5\%}/\textbf{77.8\%} & n/a  & n/a & n/a  & n/a  & 42.7\%/67.6\% \\ \hline

\end{tabular}
\end{small}
\label{tab:resnete18}
\end{table*}

\subsection{Full-Precision Pre-Training}
\label{sec:pre-training}
\begin{figure}[]
    \vspace{-0.5cm}
    \centering
    \includegraphics[width=0.9\linewidth]{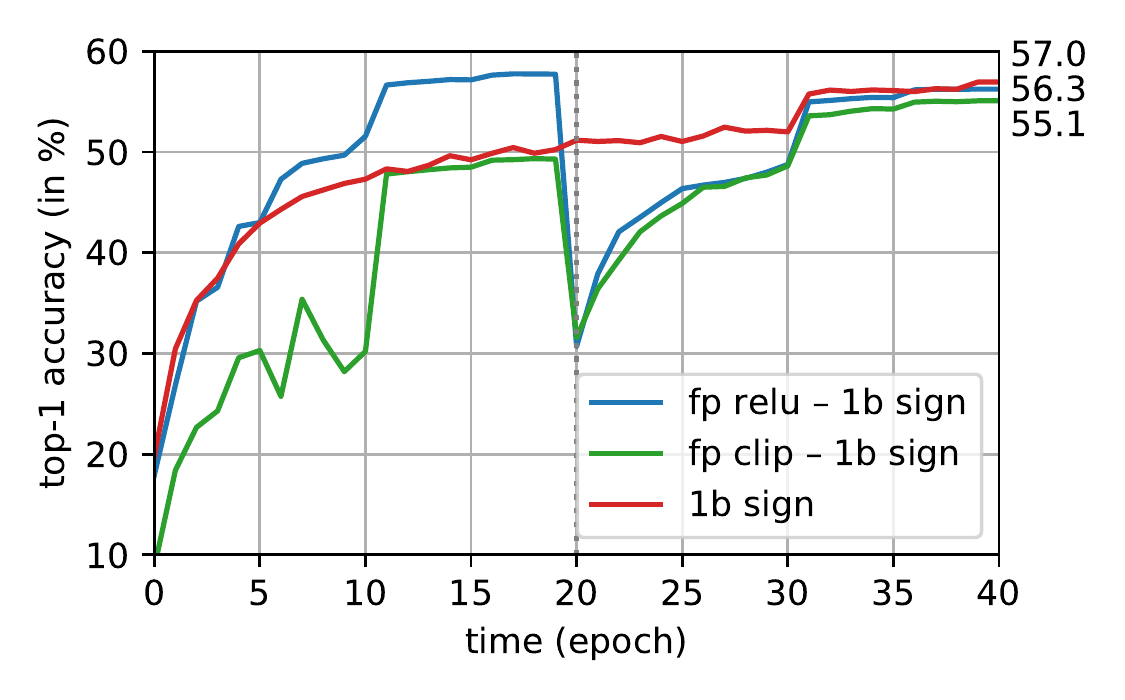}
    \vspace{-0.3cm}
    \caption{
        Top-1 validation accuracy per epoch of training binary \architecture{ResNetE18} from scratch (red, 40 epochs, Adam), from a full-precision pre-training (20 epochs, SGD) with clip (green) and ReLU (blue) as activation function.
        The degradation peak of the green and blue curve at epoch 20 depicts a heavy ``re-learning'' effect when we start fine-tuning a full-precision model to a binary one.
    }
    \label{fig:finetuning}
\vspace{-0.3cm}
\end{figure}

\noindent
Fine-tuning a full-precision model to a binary one is beneficial only if it yields better results in comparable, total training time.
We trained our binary \architecture{ResNetE18} in three different ways: fully from scratch (1), by fine-tuning a full-precision \architecture{ResNetE18} with ReLU (2) and clip (proposed by \cite{Liu_2018_ECCV}) (3) as activation function (see \autoref{fig:finetuning}).
The full-precision trainings followed the typical configuration of momentum SGD with weight decay over 20 epochs with learning rate decay of 0.1 after 10 and 15 epochs. 
For all binary trainings, we used Adam \cite{kingma2014adam} without weight decay with learning rate updates at epoch 10 and 15 for the fine-tuning and 30 and 38 for the full binary training.
Our experiment shows that clip performs worse than ReLU for fine-tuning and in general.
Additionally, the training from scratch yields a slightly better result than with pre-training.
Pre-training inherently adds complexity to the training procedure, because the different architecture of binary networks does not allow to use published ReLU models.
Thus, we advocate the avoidance of fine-tuning full-precision models.
Note that our observations are based on the involved architectures in this work.
A more comprehensive evaluation of other networks remains as future work.

\subsection{Backward Pass of the Sign Function}
\label{sec:results-approx}

\noindent
Liu \etal \cite{Liu_2018_ECCV} claim that a differentiable approximation function, called $\mathrm{approxsign}$, can be made by replacing the backward pass with
\begin{align}
    ~\frac{\partial c}{\partial r_i}=\frac{\partial c}{\partial r_o}1_{|r_i|\leq t_\mathrm{clip}}\cdot\begin{cases} 
    2-2r_i ~\text{if}~ r_i \geq 0, \\
    2+2r_i ~\text{otherwise}.
    \end{cases}
\end{align}
Since this could also benefit when training a binary network from scratch, we evaluated this in our experiments.
We compared the regular backward pass $\mathrm{sign}$ with $\mathrm{approxsign}$.
First, the results of our CIFAR-10 experiments seem to depend on whether we use scaling or not.
If we use scaling, both functions perform similarly (see \autoref{tab:cifar-scaling-downsampling-approxsign}).
Without scaling the $\mathrm{approxsign}$ function leads to less accurate models on CIFAR-10.
In our experiments on ImageNet, the performance difference between the use of the functions is minimal (see \autoref{tab:imagenet-scaling-downsampling-approxsign}). 
We conclude that applying $\mathrm{approxsign}$ instead of $\mathrm{sign}$ function seems to be specific to fine-tuning from full-precision models \cite{Liu_2018_ECCV}.
We thus don't use $\mathrm{approxsign}$ in the rest of the paper for simplicity.


\section{Proposed Approach}
\label{sec:approach}

\noindent
In this section, we present several essential design principles for training accurate BNNs from scratch.
We then practiced our design philosophy on the binary \arch{ResNetE} model, where we believe that the shortcut connections are essential for an accurate BNN.
Based on the insights learned we propose a new BNN model \arch{BinaryDenseNet} which reaches state-of-the-art accuracy without tricks.

\subsection{Golden Rules for Training Accurate BNNs}
\label{sec:golden_rules}

\noindent
As shown in \autoref{tab:resnete18}, with a standard training strategy our binary \arch{ResNetE18} model outperforms other state-of-the-art binary models by using the same network structure.
We successfully train our model from scratch by following several general design principles for BNNs, summarized as follows:
\vspace{-0.2cm}
\begin{itemize}
	\itemsep0em
	\item The core of our theory is \emph{maintaining rich information flow of the network}, which can effectively compensate the precision loss caused by quantization.

	\item Not all the well-known real-valued network architectures can be seamlessly applied for BNNs.
	The network architectures from the category \emph{compact network design} are not well suited for BNNs, since their design philosophies are mutually exclusive (eliminating redundancy $\leftrightarrow$ compensating information loss).

	\item 
	\emph{Bottleneck} design \cite{szegedy2015going} should be eliminated in your BNNs.
	We will discuss this in detail in the following paragraphs (also confirmed by \cite{Bethge2018}).

	\item
	Seriously consider using full-precision downsampling layer in your BNNs to preserve the information flow.

	\item 
	Using shortcut connections is a straightforward way to avoid bottlenecks of information flow, which is particularly essential for BNNs.	
	
  \item
  To overcome bottlenecks of information flow, we should appropriately increase the network width (the dimension of feature maps) while going deeper (as \eg, see \arch{BinaryDenseNet37/37-dilated/45} in \autoref{tab:imagenet-full-downs}).
  However, this may introduce additional computational costs.

	\item
	The previously proposed complex training strategies, as \eg. scaling factors, $\mathrm{approxsign}$ function, FP pre-training are \emph{not} necessary to reach state-of-the-art performance when training a binary model directly from scratch.
\end{itemize}

Before thinking about model architectures, we must consider the main drawbacks of BNNs.
First of all, the information density is theoretically 32 times lower, compared to full-precision networks.
Research suggests, that the difference between 32 bits and 8 bits seems to be minimal and 8-bit networks can achieve almost identical accuracy as full-precision networks \cite{Han2015}.
However, when decreasing bit-width to four or even one bit (binary), the accuracy drops significantly \cite{Courbariaux2016,Zhou2016}.
Therefore, the precision loss needs to be alleviated through other techniques, for example by increasing information flow through the network.
We further describe three main methods in detail, which help to preserve information despite binarization of the model:

First, a binary model should use as many shortcut connections as possible in the network.
These connections allow layers later in the network to access information gained in earlier layers despite of precision loss through binarization.
Furthermore, this means that increasing the number of connections between layers should lead to better model performance, especially for binary networks.

Secondly, network architectures including bottlenecks are always a challenge to adopt.
The bottleneck design reduces the number of filters and values significantly between the layers, resulting in less information flow through BNNs.
Therefore we hypothesize that either we need to eliminate the bottlenecks or at least increase the number of filters in these bottleneck parts for BNNs to achieve best results.

The third way to preserve information comes from replacing certain crucial layers in a binary network with full precision layers.
The reasoning is as follows:
If layers that do not have a shortcut connection are binarized, the information lost (due to binarization) can not be recovered in subsequent layers of the network.
This affects the first (convolutional) layer and the last layer (a fully connected layer which has a number of output neurons equal to the number of classes), as learned from previous work \cite{Rastegari2016,Zhou2016,Liu_2018_ECCV,Wan_2018_ECCV,Courbariaux2016}.
These layers generate the initial information for the network or consume the final information for the prediction, respectively.
Therefore, full-precision layers for the first and the final layer are always applied previously.
Another crucial part of deep networks is the \emph{downsampling convolution} which converts all previously collected information of the network to smaller feature maps with more channels (this convolution often has stride two and output channels equal to twice the number of input channels).
Any information lost in this downsampling process is effectively no longer available.
Therefore, it should always be considered whether these downsampling layers should be in full-precision, even though it slightly increases model size and number of operations.

\subsection{ResNetE}
\label{sec:resnetE}

\begin{figure}[]
\captionsetup[subfigure]{justification=centering}
\begin{center}
\begin{subfigure}[t]{0.32\linewidth}
   \centering
   \includegraphics[width=0.915\linewidth]{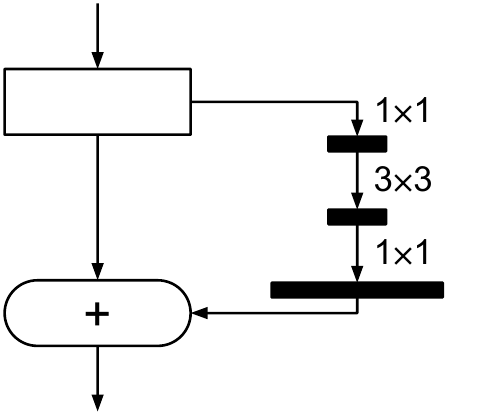}
   \caption{ResNet \\ (bottleneck)}
   \label{fig:netblocks-resnet-bottleneck}
\end{subfigure}
\begin{subfigure}[t]{0.32\linewidth}
   \centering
   \includegraphics[width=0.915\linewidth]{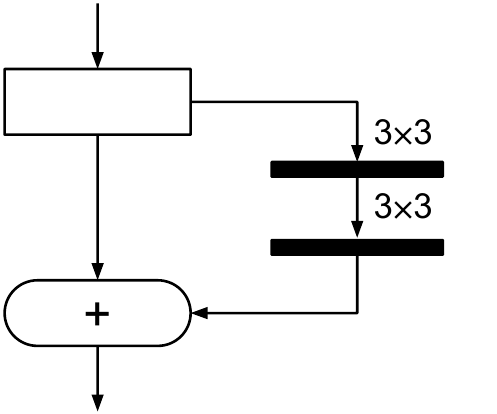}
   \caption{ResNet \\ (no bottleneck)}
   \label{fig:netblocks-resnet}
\end{subfigure}
\begin{subfigure}[t]{0.32\linewidth}
   \centering
   \includegraphics[width=0.915\linewidth]{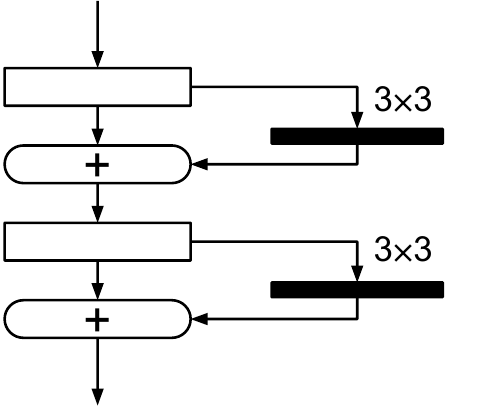}
   \caption{ResNetE \\ (added shortcut)}
   \label{fig:netblocks-resnete}
\end{subfigure}
\begin{subfigure}[t]{0.32\linewidth}
   \centering
   \includegraphics[width=0.99\linewidth]{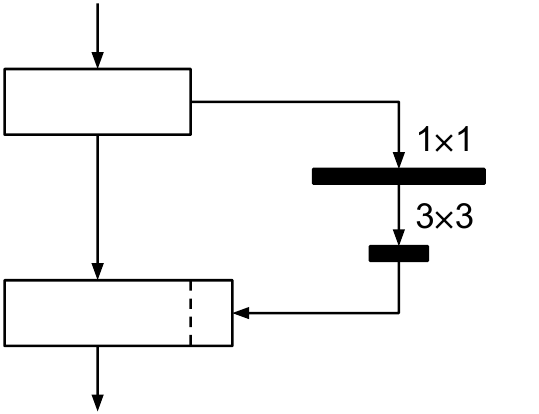}
   \caption{DenseNet \\ (bottleneck)}
   \label{fig:netblocks-densenet-bottleneck}
\end{subfigure}
\begin{subfigure}[t]{0.32\linewidth}
   \centering
   \includegraphics[width=0.99\linewidth]{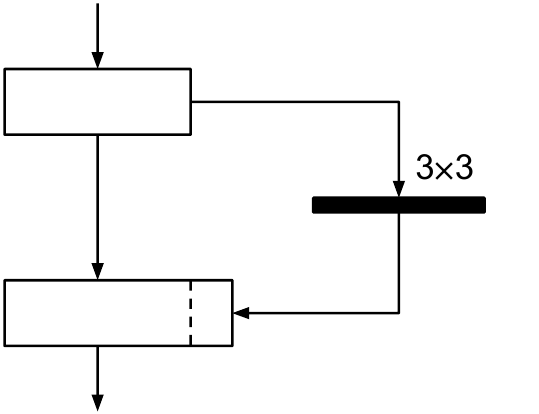}
   \caption{DenseNet \\ (no bottleneck)}
   \label{fig:netblocks-densenet}
\end{subfigure}
\begin{subfigure}[t]{0.32\linewidth}
   \centering
   \includegraphics[width=0.99\linewidth]{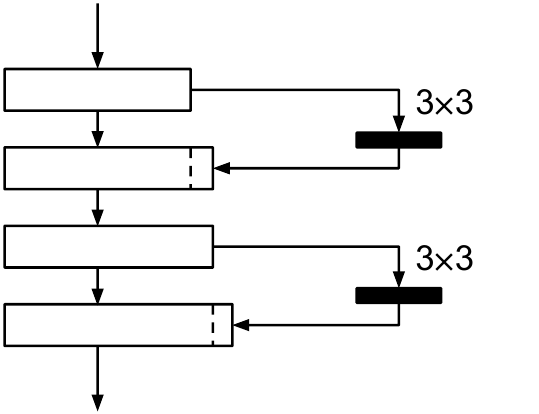}
   \caption{BinaryDenseNet}
   \label{fig:netblocks-densenet-split}
\end{subfigure}
\end{center}
\vspace{-0.3cm}
\caption{
   A single building block of different network architectures (the length of bold black lines represents the number of filters).
    (\subref{fig:netblocks-resnet-bottleneck}) The original \architecture{ResNet} design features a bottleneck architecture.
    A low number of filters reduces information capacity for BNNs.
    (\subref{fig:netblocks-resnet}) A variation of the \architecture{ResNet} without the bottleneck design.
    The number of filters is increased, but with only two convolutions instead of three.
    (\subref{fig:netblocks-resnete}) The \architecture{ResNet} architecture with an additional shortcut, first introduced in \cite{Liu_2018_ECCV}.
    (\subref{fig:netblocks-densenet-bottleneck}) The original \architecture{DenseNet} design with a bottleneck in the second convolution operation.
    (\subref{fig:netblocks-densenet}) The \architecture{DenseNet} design without a bottleneck.
    The two convolution operations are replaced by one $3\times3$ convolution.
    (\subref{fig:netblocks-densenet-split}) Our suggested change to a \architecture{DenseNet} where a convolution with N filters is replaced by two layers with $\frac{N}{2}$ filters each.
}
\label{fig:netblocks}
\end{figure}

\noindent
\architecture{ResNet} combines the information of all previous layers with shortcut connections.
This is done by adding the input of a block to its output with an identity connection. 
As suggested in the previous section, we remove the bottleneck of a \arch{ResNet} block by replacing the three convolution layers (kernel sizes 1, 3, 1) of a regular \arch{ResNet} block with two $3\times3$ convolution layers with a higher number of filters (see \autoref{fig:netblocks-resnet-bottleneck}, \subref{fig:netblocks-resnet}).
We subsequently increase the number of connections by reducing the block size from two convolutions per block to one convolution per block, as inspired by \cite{Liu_2018_ECCV}.
This leads to twice the amount of shortcuts, as there are as many shortcuts as blocks, if the amount of layers is kept the same (see \autoref{fig:netblocks-resnete}).
However, \cite{Liu_2018_ECCV} also incorporates other changes to the \arch{ResNet} architecture. 
Therefore we call this specific change in the block design \architecture{ResNetE} (short for \textbf{E}xtra shortcut).
The second change is using the full-precision downsampling convolution layer (see \autoref{fig:transition-resnet}).
In the following we conduct an ablation study for testing the exact accuracy gain and the impact of the model size.

We evaluated the difference between using binary and full-precision downsampling layers, which has been often ignored in the literature.
First, we examine the results of binary \arch{ResNetE18} on CIFAR-10.
Using full-precision downsampling over binary leads to an accuracy gain between 0.2\% and 1.2\% (see \autoref{tab:cifar-scaling-downsampling-approxsign}).
However, the model size also increases from 1.39 MB to 2.03 MB, which is arguably too much for this minor increase of accuracy.
Our results show a significant difference on ImageNet (see \autoref{tab:imagenet-scaling-downsampling-approxsign}).
The accuracy increases by 3\% when using full-precision downsampling.
Similar to CIFAR-10, the model size increases by 0.64 MB, in this case from 3.36 MB to 4.0 MB.
The larger base model size makes the relative model size difference lower and provides a stronger argument for this trade-off.
We conclude that the increase in accuracy is significant, especially for ImageNet. 

Inspired by the achievement of binary \arch{ResNetE}, we naturally further explored the \arch{DenseNet} architecture, which is supposed to benefit even more from the densely connected layer design.

\subsection{BinaryDenseNet}
\label{sec:binarydensenet}

\begin{figure}[]
\captionsetup[subfigure]{justification=centering}
\begin{center}
\begin{subfigure}[t]{0.36\linewidth}
   \centering
   \includegraphics[width=0.95\linewidth]{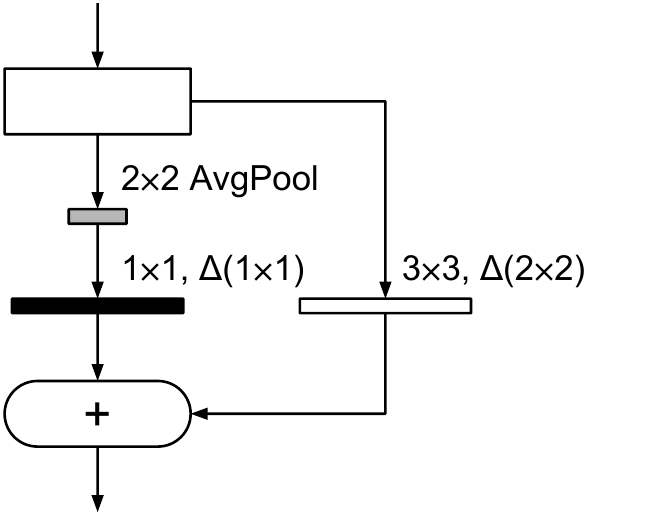}
   \caption{ResNet}
   \label{fig:transition-resnet}
\end{subfigure}
\begin{subfigure}[t]{0.27\linewidth}
   \centering
   \includegraphics[width=0.95\linewidth]{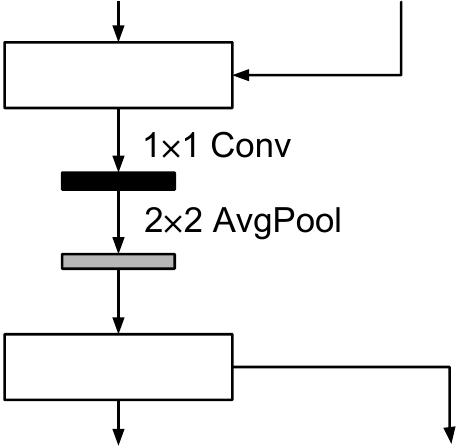}
   \caption{DenseNet}
   \label{fig:transition-densenet}
\end{subfigure}
\begin{subfigure}[t]{0.27\linewidth}
   \centering
   \includegraphics[width=0.95\linewidth]{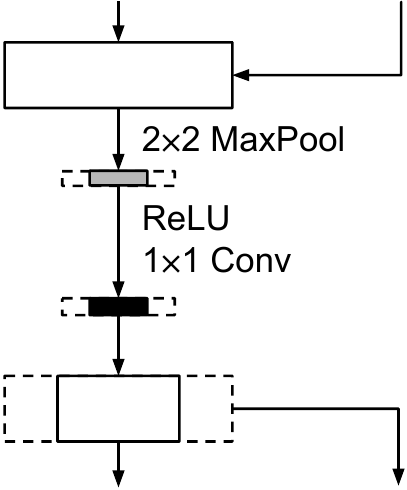}
   \caption{BinaryDenseNet}
   \label{fig:transition-binary-densenet}
\end{subfigure}
\end{center}
\vspace{-0.3cm}
\caption{
The downsampling layers of \arch{ResNet}, \arch{DenseNet} and \arch{BinaryDenseNet}.
The bold black lines mark the downsampling layers which can be replaced with FP layers.
If we use FP downsampling in a \arch{BinaryDenseNet}, we increase the reduction rate to reduce the number of channels (the dashed lines depict the number of channels without reduction).
We also swap the position of pooling and Conv layer that effectively reduces the number of MACs. 
}
\label{fig:downsampling}
\vspace{-0.3cm}
\end{figure}
%

\begin{table}[]
\caption{
The difference of performance for different \arch{BinaryDenseNet} models when using different downsampling methods evaluated on ImageNet.
}
\begin{tabular}{|l|l|l|l|}
\hline
\lmrcell{Blocks, \\ growth-rate} & \mrcell{Model \\ size \\ (binary)} & \mrcell{Downsampl. \\ convolution, \\ reduction} & \mrcell{Accuracy \\ Top1/Top5}      \\ \hline
\multirow{2}{*}{16, 128}           & 3.39 MB    & binary, low   & 52.7\%/75.7\%                   \\ \cline{2-4}
                                        & 3.03 MB    & FP, high  & \textbf{55.9\%}/\textbf{78.5\%} \\ \hline
\multirow{2}{*}{32, 64}            & 3.45 MB    & binary, low   & 54.3\%/77.3\%                   \\ \cline{2-4}
                                        & 3.08 MB    & FP, high  & \textbf{57.1\%}/\textbf{80.0\%} \\ \hline
\end{tabular}
\label{tab:densenet-downsampling}
\end{table}

\noindent
\architecture{DenseNets} \cite{Huang2016} apply shortcut connections that, contrary to \arch{ResNet}, concatenate the input of a block to its output (see \autoref{fig:netblocks-densenet-bottleneck}, \subref{fig:netblocks-resnet}).
Therefore, new information gained in one layer can be reused throughout the entire depth of the network.
We believe this is a significant characteristic for maintaining information flow.
Thus, we construct a novel BNN architecture: \arch{BinaryDenseNet}.

The bottleneck design and transition layers of the original \arch{DenseNet} effectively keep the network at a smaller total size, even though the concatenation adds new information into the network every layer.
However, as previously mentioned, we have to eliminate bottlenecks for BNNs.
The bottleneck design can be modified by replacing the two convolution layers (kernel sizes 1 and 3) with one $3\times3$ convolution (see \autoref{fig:netblocks-densenet-bottleneck}, \subref{fig:netblocks-densenet}). 
However, our experiments showed that \architecture{DenseNet} architecture does not achieve satisfactory performance, even after this change.
This is due to the limited representation capacity of binary layers. 
There are different ways to increase the capacity.
We can increase the growth rate parameter $k$, which is the number of newly concatenated features from each layer.
We can also use a larger number of blocks.
Both individual approaches add roughly the same amount of parameters to the network.
To keep the number of parameters equal for a given \architecture{BinaryDenseNet} we can halve the growth rate and double the number of blocks at the same time (see \autoref{fig:netblocks-densenet-split}) or vice versa.
We assume that in this case increasing the number of blocks should provide better results compared to increasing the growth rate. 
This assumption is derived from our hypothesis: favoring an increased number of connections over simply adding weights.

Another characteristic difference of \arch{BinaryDenseNet} compared to binary \arch{ResNetE} is that the downsampling layer reduces the number of channels.
To preserve information flow in these parts of the network we found two options:
On the one hand, we can use a full-precision downsampling layer, similarly to binary \arch{ResNetE}.
Since the full-precision layer preserves more information, we can use higher reduction rate for downsampling layers.
To reduce the number of MACs, we modify the transition block by swapping the position of pooling and convolution layers.
We use \emph{MaxPool}$\rightarrow$\emph{ReLU}$\rightarrow$1$\times$1-\emph{Conv} instead of 1$\times$1-\emph{Conv}$\rightarrow$\emph{AvgPool} in the transition block (see \autoref{fig:transition-binary-densenet}, \subref{fig:transition-densenet}).
On the other hand, we can use a binary downsampling conv-layer instead of a full-precision layer with a lower reduction rate, or even no reduction at all.
We coupled the decision whether to use a binary or a full-precision downsampling convolution with the choice of reduction rate.
The two variants we compare in our experiments (see \autoref{sec:ablation-binaryDensenet}) are thus called \emph{full-precision downsampling with high reduction} (halve the number of channels in all transition layers) and \emph{binary downsampling with low reduction} (no reduction in the first transition, divide number of channels by 1.4 in the second and third transition).

\subsubsection{Experiment}
\label{sec:ablation-binaryDensenet}
\noindent
\textbf{Downsampling Layers.}
In the following we present our evaluation results of a \arch{BinaryDenseNet} when using a full-precision downsampling with high reduction over a binary downsampling with low reduction.
The results of a \arch{BinaryDenseNet21} with growth rate 128 for CIFAR-10 result show an accuracy increase of 2.7\% from 87.6\% to 90.3\%.
The model size increases from 673 KB to 1.49 MB.
This is an arguably sharp increase in model size, but the model is still smaller than a comparable binary \arch{ResNet18} with a much higher accuracy.
The results of two \arch{BinaryDenseNet} architectures (16 and 32 blocks combined with 128 and 64 growth rate respectively) for ImageNet show an increase of accuracy ranging from 2.8\% to 3.2\% (see \autoref{tab:densenet-downsampling}).
Further, because of the higher reduction rate, the model size decreases by 0.36 MB at the same time.
This shows a higher effectiveness and efficiency of using a FP downsampling layer for a \arch{BinaryDenseNet} compared to a binary \arch{ResNet}.

%
\begin{table}[]
\caption{
The accuracy of different \arch{BinaryDenseNet} models by successively splitting blocks evaluated on ImageNet.
As the number of connections increases, the model size (and number of binary operations) changes marginally, but the accuracy increases significantly.
}
\begin{tabular}{|l|l|l|l|}
\hline
\lmrcell{Blocks} & \mrcell{Growth-\\ rate} & \mrcell{Model size \\ (binary)} & \mrcell{Accuracy \\ Top1/Top5}      \\ \hline
8   & 256         & 3.31 MB    & 50.2\%/73.7\% \\ \hline
16  & 128         & 3.39 MB    & 52.7\%/75.7\% \\ \hline
32  & 64          & 3.45 MB    &\textbf{55.5\%}/\textbf{78.1\%} \\ \hline
\end{tabular}
\label{tab:densenet-growth-rate-vs-layers}
\vspace{-0.3cm}
\end{table}
\noindent
\textbf{Splitting Layers.}
\label{sec:results-split-densenet}
We tested our proposed architecture change (see \autoref{fig:netblocks-densenet-split}) by comparing \arch{BinaryDenseNet} models with varying growth rates and number of blocks (and thus layers).
The results show, that increasing the number of connections by adding more layers over simply increasing growth rate increases accuracy in an efficient way (see \autoref{tab:densenet-growth-rate-vs-layers}).
Doubling the number of blocks and halving the growth rate leads to an accuracy gain ranging from 2.5\% to 2.8\%.
Since the training of a very deep \arch{BinaryDenseNet} becomes slow (it is less of a problem during inference, since no additional memory is needed during inference for storing some intermediate results), we have not trained even more highly connected models, but highly suspect that this would increase accuracy even further.
The total model size slightly increases, since every second half of a split block has slightly more inputs compared to those of a double-sized normal block.
In conclusion, our technique of increasing number of connections is highly effective and size-efficient for a \arch{BinaryDenseNet}.

\begin{table}[]
\caption{Comparison of our \arch{BinaryDenseNet} to state-of-the-art 1-bit CNN models on ImageNet.}
\vspace{-0.3cm}
\begin{small}
\begin{center}
\begin{tabular}{|l|l|l|l|}
\hline
 \thead{Model\\size} &  Method  &  \thead{Top-1/Top-5\\accuracy}      \\ \hline

\multirow{5}{*}{$\sim$4.0MB} & XNOR-ResNet18 \cite{Rastegari2016} & 51.2\%/73.2\%     \\ \cline{2-3}
 							 & TBN-ResNet18 \cite{Wan_2018_ECCV} & 55.6\%/74.2\%      \\ \cline{2-3}
 							 & Bi-Real-ResNet18 \cite{Liu_2018_ECCV} & 56.4\%/79.5\%  \\ \cline{2-3}
 							 & \arch{BinaryResNetE18} & 58.1\%/80.6\%       \\ \cline{2-3}
 							 & \arch{BinaryDenseNet28} & \textbf{60.7\%}/\textbf{82.4\%}   \\ \hline
\multirow{4}{*}{$\sim$5.1MB} & TBN-ResNet34 \cite{Wan_2018_ECCV} & 58.2\%/81.0\%         \\ \cline{2-3} 			
 							 & Bi-Real-ResNet34 \cite{Liu_2018_ECCV} & 62.2\%/83.9\%     \\ \cline{2-3}
 							 & \arch{BinaryDenseNet37} & 62.5\%/83.9\%       \\ \cline{2-3}
 							 & \arch{BinaryDenseNet37-dilated}$^*$ & \textbf{63.7\%}/\textbf{84.7\%}   \\ \hline
 		7.4MB				 & \arch{BinaryDenseNet45}  & 63.7\%/84.8\%   \\ \hline
 		46.8MB				 & Full-precision ResNet18 & 69.3\%/89.2\%   \\ 		
 		249MB				 & Full-precision AlexNet  & 56.6\%/80.2\%   \\ \hline 		 		
\end{tabular}
\end{center}
\vspace{-0.3cm}
{\small $^*$ \arch{BinaryDenseNet37-dilated} is slightly different to other models as it applies dilated convolution kernels, while the spatial dimention of the feature maps are unchanged in the 2nd, 3rd and 4th stage that enables a broader information flow.}
\end{small}
\vspace{-0.3cm}
\label{tab:imagenet-full-downs}
\end{table}
%


\section{Main Results}
\label{sec:main_results}
\noindent
In this section, we report our main experimental results on image classification and object detection using \arch{BinaryDenseNet}.
We further report the computation cost in comparison with other quantization methods.
Our implementation is based on the \emph{BMXNet} framework first presented by Yang \etal \cite{HPI_xnor}.
Our models are trained from scratch using a standard training strategy.
Due to space limitations, more details of the experiment can be found in the supplementary materials.

\noindent
\textbf{Image Classification.}
To evaluate the classification accuracy, we report our results on ImageNet~\cite{imagenet_cvpr09}.
\autoref{tab:imagenet-full-downs} shows the comparison result of our \arch{BinaryDenseNet} to state-of-the-art BNNs with different sizes.
For this comparison, we chose growth and reduction rates for \arch{BinaryDenseNet} models to match the model size and complexity of the corresponding binary \arch{ResNet} architectures as closely as possible.
Our results show that \arch{BinaryDenseNet} surpass all the existing 1-bit CNNs with noticeable margin.
Particularly, \arch{BinaryDenseNet28} with 60.7\% top-1 accuracy, is better than our binary \arch{ResNetE18}, and achieves up to 18.6\% and 7.6\% relative improvement over the well-known XNOR-Network and the current state-of-the-art Bi-Real Net, even though they use a more complex training strategy and additional techniques, \eg, custom gradients and a scaling variant.

\noindent
\textbf{Preliminary Result on Object Detection.}
We adopted the off-the-shelf toolbox Gluon-CV \cite{he2018bag} for the object detection experiment.
We change the base model of the adopted SSD architecture \cite{LiuAESRFB16} to \arch{BinaryDenseNet} and
train our models on the combination of PASCAL VOC2007 trainval and VOC2012 trainval, and test on VOC2007 test set \cite{Everingham:2010}.
\autoref{tab:voc-detection} illustrates the results of binary SSD as well as some FP detection models \cite{LiuAESRFB16,Faster-rcnn,YOLOv1}.

\begin{table}[]
\caption{Object detection performance (in mAP) of our \arch{BinaryDenseNet37/45} and other BNNs on VOC2007 test set.}
\vspace{-0.5cm}
\begin{small}
\begin{center}
\begin{tabular}{|l|c|c|c|}
\hline
 Method & \thead{Ours$^\dag$\\37/45}  &  \thead{TBN$^*$\\ResNet34} &  \thead{XNOR-Net$^*$\\ResNet34}  \\ \hline
 Binary SSD  & 	\textbf{66.4/68.2} &  59.5  &  55.1    \\ \hline 
 \thead{Full-precision\\SSD512/faster rcnn/yolo}  & \multicolumn{3}{c|}{76.8/73.2/66.4} \\ \hline 			  
\end{tabular}
\end{center}
\vspace{-0.3cm}
{\small $^*$ SSD300 result read from \cite{Wan_2018_ECCV}, $^\dag$ SSD512 result}
\end{small}
\vspace{-0.3cm}
\label{tab:voc-detection}
\end{table}

\noindent
\textbf{Efficiency Analysis.}
\begin{figure}[]
    \centering
    \includegraphics[width=0.8\linewidth]{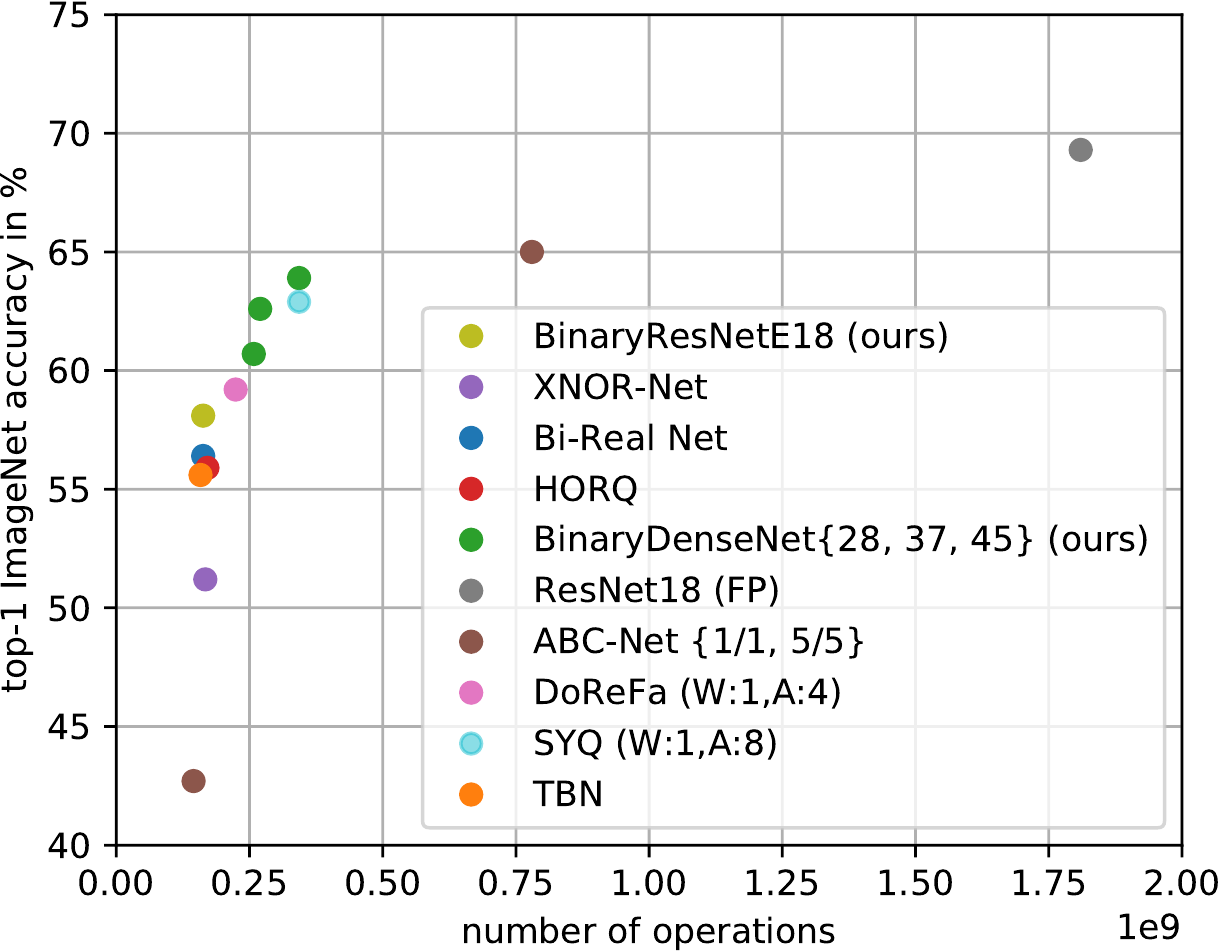}
    \vspace{-0.2cm}
    \caption{
       The trade-off of top-1 validation accuracy on ImageNet and number of operations.
       All the binary/quantized models are based on \arch{ResNet18} except \arch{BinaryDenseNet}.
    }
    \label{fig:computation-flops}
\vspace{-0.4cm}
\end{figure}
For this analysis, we adopted the same calculation method as \cite{Liu_2018_ECCV}.
\autoref{fig:computation-flops} shows that our binary \arch{ResNetE18} demonstrates higher accuracy with the same computational complexity compared to other BNNs, and \arch{BinaryDenseNet28/37/45} achieve significant accuracy improvement with only small additional computation overhead.
For a more challenging comparison we include models with 1-bit weight and multi-bits activations: DoReFa-Net (w:1, a:4) \cite{Zhou2016}	and SYQ (w:1, a:8) \cite{faraone2018syq}, and a model with multiple weight and multiple activation bases: ABC-Net \{5/5\}. 
Overall, our \arch{BinaryDenseNet} models show superior performance while measuring both accuracy and computational efficiency.

In closing, although the task is still arduous, we hope the ideas and results of this paper will provide new potential directions for the future development of BNNs.

{\small
\bibliographystyle{ieee}
\bibliography{bmxnet2}
}

\end{document}